\documentclass[letterpaper, 10 pt, conference， anonymous]{ieeeconf}  % Comment this line out if you need a4paper
\usepackage{xcolor}

\IEEEoverridecommandlockouts                              % This command is only needed if 
\usepackage{hyperref}
\usepackage{graphicx}
\usepackage{amsfonts}
\usepackage{algorithm}
\usepackage{algpseudocode}

\usepackage{amsmath} % assumes amsmath package installed

\usepackage{booktabs}   % 用于生成专业的三线表
\usepackage{multirow}

\title{\LARGE \bf
Offline Discovery of Interpretable Skills from Multi-Task Trajectories
}

\author{Chongyu Zhu$^{1\dagger*}$, Mithun Vanniasinghe$^{2\dagger}$, Jiayu Chen$^{3}$, and Chi-Guhn Lee$^{1}$%
\thanks{\textdagger Equal contribution.}%
\thanks{$^{1}$C. Zhu and C.-G. Lee are with the Department of Mechanical and Industrial Engineering, and the Operation Research and Reinforcement Learning (DORL) Lab, University of Toronto, Toronto, ON, Canada.}%
\thanks{$^{2}$M. Vanniasinghe is with the University of Toronto Institute for Aerospace Studies (UTIAS), Toronto, ON, Canada.}%
\thanks{$^{3}$J. Chen is with the Agentic Intelligence Lab, The University of Hong Kong, Hong Kong SAR, China.}%
\thanks{*Corresponding author: {\tt\small cyzhu@mie.utoronto.ca}}%
}

\begin{document}

\maketitle
\thispagestyle{empty}
\pagestyle{empty}

%%%%%%%%%%%%%%%%%%%%%%%%%%%%%%%%%%%%%%%%%%%%%%%%%%%%%%%%%%%%%%%%%%%%%%%%%%%%%%%%
\begin{abstract}

Hierarchical Imitation Learning is a powerful paradigm for acquiring complex robot behaviors from demonstrations. A central challenge, however, lies in discovering reusable skills from long-horizon, multi-task offline data, especially when the data lacks explicit rewards or subtask annotations. In this work, we introduce LOKI, a three-stage end-to-end learning framework designed for offline skill discovery and hierarchical imitation. The framework commences with a two-stage, weakly supervised skill discovery process: Stage one performs coarse, task-aware macro-segmentation by employing an alignment-enforced Vector Quantized VAE guided by weak task labels. Stage two then refines these segments at a micro-level using a self-supervised sequential model, followed by an iterative clustering process to consolidate skill boundaries. The third stage then leverages these precise boundaries to construct a hierarchical policy within an option-based framework—complete with a learned termination condition $\beta$ for explicit skill switching. LOKI achieves high success rates on the challenging D4RL Kitchen benchmark and outperforms standard HIL baselines. Furthermore, we demonstrate that the discovered skills are semantically meaningful, aligning with human intuition, and exhibit compositionality by successfully sequencing them to solve a novel, unseen task. 
% Our code will be made publicly available after acceptance.
The code is available at: \href{https://github.com/cyzhu-hiter/LOKI}{https://github.com/cyzhu-hiter/LOKI}

\end{abstract}

%%%%%%%%%%%%%%%%%%%%%%%%%%%%%%%%%%%%%%%%%%%%%%%%%%%%%%%%%%%%%%%%%%%%%%%%%%%%%%%%
\section{INTRODUCTION}
\label{sec:introduction}
% \chongyu{Motivation}
The development of intelligent robotic systems capable of performing complex, long-horizon tasks in diverse and unstructured environments \cite{intelligence2025pi05} stands as a central goal of modern artificial intelligence and robotics research. Traditional methods, whether relying on meticulous manual programming and rule-based control or the challenging design of dense reward functions in reinforcement learning, often encounter significant hurdles in scalability, adaptability, and debuggability when confronted with such intricate tasks \cite{xue2025fullmpc, zare2024ilsurvey}. In response, Imitation Learning (IL), which empowers robots to acquire behaviors directly from human demonstrations, has emerged as a particularly promising paradigm. IL effectively circumvents the complexities of explicit policy engineering and sparse reward signals, making it highly valuable for tasks where direct programming is infeasible or costly \cite{zare2024ilsurvey, gupta2019relaypolicylearning}. Building upon this foundation, Hierarchical Imitation Learning (HIL) further advances the field by decomposing complex tasks into manageable subtasks, thereby simplifying control and facilitating the learning of long-horizon behaviors. HIL has achieved notable success in various domains, from sophisticated robotic manipulation and assembly to robust autonomous navigation, and is increasingly favored for deploying modular policies on resource-constrained computational units rather than comprehensive general models \cite{zhong2025vlasurvey, venkatraman2023reasoningdiffusion, hierdriving}.

% \chongyu{Problem Statement}
However, scaling IL to truly complex, multi-task scenarios presents significant challenges. Real-world human demonstrations, while rich in information, are often collected offline, lack explicit subtask labels, and feature long, unsegmented sequences of actions that span multiple distinct behaviors. Discovering a set of reusable, semantically meaningful skills from such raw, unannotated offline multi-task data remains a formidable obstacle \cite{zare2024ilsurvey}. Existing methods often struggle with several key issues: (1) Ambiguity in Skill Boundaries: Accurately segmenting continuous trajectories into discrete, purposeful skills without explicit supervision is difficult, leading to sub-optimal skill definitions \cite{mao2024dexskills}. (2) Lack of Reusability and Interpretability: Learned skills may be highly task-specific or fail to align with human conceptualizations, hindering their transferability to novel tasks or human-robot collaboration \cite{zare2024ilsurvey, wen2024diffusionreasoningvla}. (3) Scalability to Multi-Task Settings: Many approaches are designed for single-task settings or require extensive task-specific engineering, limiting their applicability in diverse, multi-task environments where generalizable skills are paramount \cite{chen2023multi}. While HIL offers a compelling framework to manage long-horizon tasks by composing simpler skills, its effectiveness fundamentally hinges on the quality and semantic coherence of these underlying discovered skills \cite{byrne1998learningbyil}.

% \chongyu{Key Insight \& Our Approach}
This work addresses these critical gaps by proposing Learning Offline Key-skills via Imitation (LOKI) method LOKI, a novel three-stage, end-to-end learning framework for robust skill discovery and hierarchical imitation from long-horizon, multi-task offline data. The design is motivated by a two-stage skill discovery philosophy that is both effective and aligns with human intuition, as conceptualized in Fig. \ref{fig:two_stage_skill_learning_diagram}. Inspired by Vector Quantized VAEs (VQ-VAE) \cite{van2017neuralvqvae}, our framework first achieves a General Skill Decomposition using a novel alignment-enforced VQ-VAE (EVQ-VAE). The resulting segmentation is then refined through a Task Skill Alignment clustering process. This approach enables robots to learn robust, interpretable, and composable skill sets directly from raw demonstrations, paving the way for more capable agentic robotic systems.
Our main contributions are:

\begin{figure}[thpb]
  \centering
  \includegraphics[width=1\linewidth]{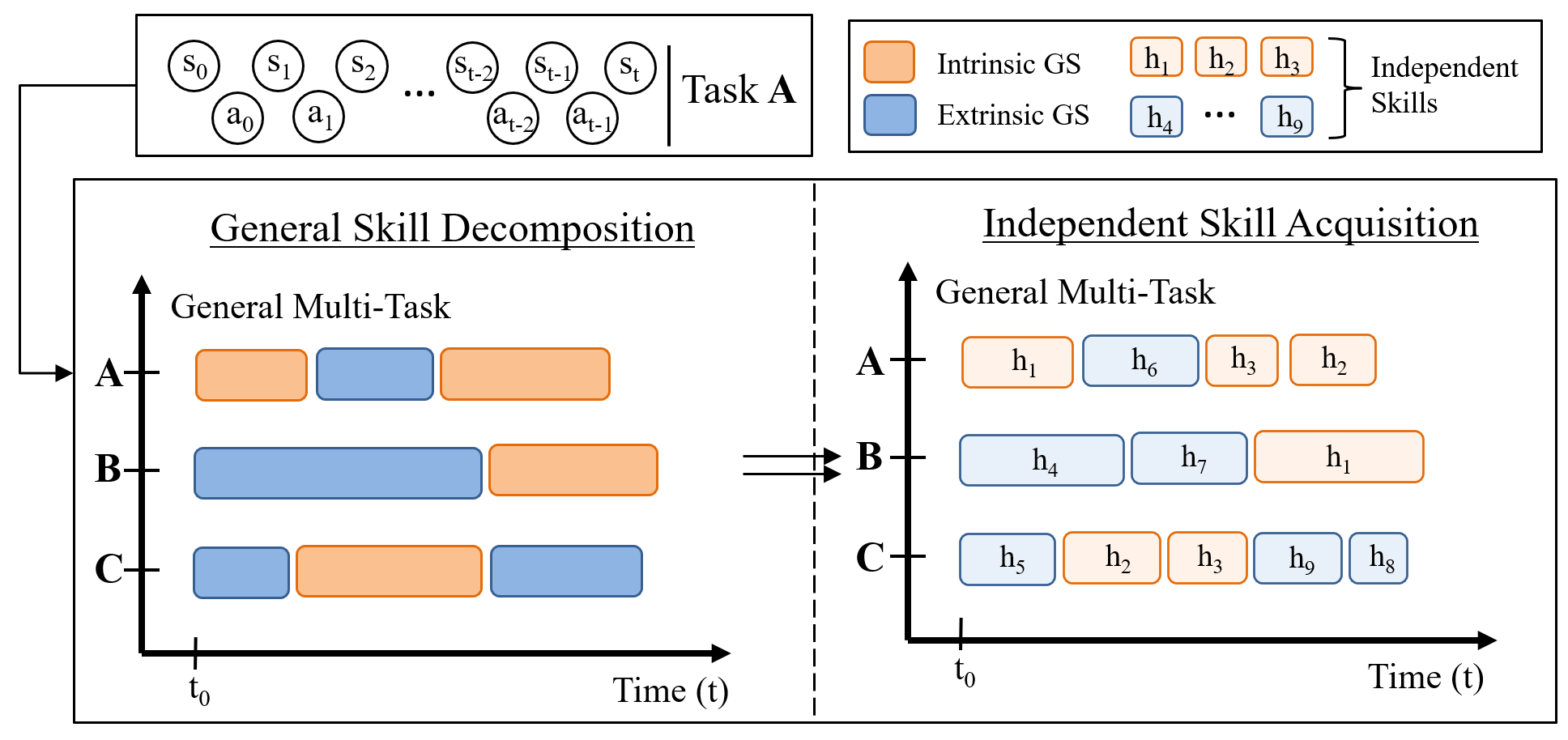}
  \caption{Two-Stage Skill Discovery: From long-horizon multi-task trajectories to reusable independent skills. While traditional pure skill discovery through prior knowledge \cite{venkatraman2023reasoningdiffusion} or information theory \cite{eysenbach2018diversity} on this problem often struggles, our approach first processes a trajectory into task-related extrinsic General Skills(GS) and task-unrelated intrinsic GS. Subsequently, these are decomposed into granular, human-knowledge-aligned Independent Skills through weak-supervision alignment. This methodology enables robust skill discovery and facilitates flexible skill recomposition for novel tasks.}
  \label{fig:two_stage_skill_learning_diagram}
\end{figure}

\begin{itemize}
    \item We propose LOKI, a novel three-stage, weakly supervised framework that discovers skills from long-horizon, multi-task offline data and enables the construction of an executable, option-based hierarchical policy.
    
    \item We introduce EVQ-VAE, a novel architecture that leverages weak task labels for coarse-grained, task-aware macro-segmentation.
    
    \item We propose a weakly supervised micro-segmentation method to distinguish conservative skills.
    
    \item We demonstrate the effectiveness of our framework, and the discovered skills are highly interpretable, consistent and composable to solve novel tasks.
\end{itemize}

\section{Related Works}
\label{sec:related_works}

\subsection{Skill Discovery}
Skill discovery aims to extract a set of reusable and meaningful behavioral primitives from unstructured trajectory data, which is a cornerstone for enabling hierarchical control and long-horizon task planning in robotics. The field can be broadly divided into online and offline approaches. Online methods typically learn skills through direct interaction with the environment, often by maximizing mutual information between states and a latent skill variable to encourage diverse and predictable behaviors \cite{eysenbach2018diversity}. Other online approaches focus on learning skills that reach diverse areas of the state space, emphasizing reachability and exploration \cite{sharma2019dynamics,DBLP:conf/nips/ChenAL23}. While powerful, these methods require extensive and often costly online interaction, which is not always feasible in real-world robotics.

In contrast, offline skill discovery operates on a fixed, pre-collected dataset of trajectories. This setting is more practical but also more challenging, as the agent cannot perform trial-and-error to self-correct and is entirely dependent on the quality and coverage of the provided data \cite{levine2020offlinechallenge}. Methodologically, offline approaches often resemble unsupervised clustering. One prominent line of work utilizes discrete generative models, such as Vector Quantized Variational Autoencoders (VQ-VAE), to learn a discrete latent space that implicitly corresponds to different skills \cite{van2017neuralvqvae, ajay2020opal}. Another approach treats skill discovery as a trajectory segmentation problem, framing trajectories as time-series data and seeking to identify change points that signify transitions between behaviors \cite{salimpour2022selfchangepointdetection}. 

\subsection{Hierarchical Imitation Learning}
HIL is an effective paradigm for solving long-horizon tasks by decomposing a complex control problem into a two-level hierarchy: a high-level policy that makes decisions in a temporally abstract manner, and a low-level policy that executes these decisions as concrete actions \cite{zare2024ilsurvey}. It is a natural framework to leverage the reusable skills identified through skill discovery. The low-level policy is typically a skill-conditioned controller, which takes the current state and a latent code from the high-level policy as input to generate actions. It is often trained via behavior cloning \cite{behaviorcloning} and concludes either upon reaching a fixed timestep or upon receiving a termination signal \cite{SUTTON1999semimdp}.

The primary focus of HIL research often lies in the design and learning of the high-level policy. One common approach is to train a neural network as a dynamic meta-controller that learns to select the optimal skill based on the current state \cite{zare2024ilsurvey}. While flexible, this approach can be complex and sample-inefficient. An alternative is to use more structured high-level policies, such as task graphs or fixed skill sequences, which can be derived from prior knowledge or extracted directly from demonstration data \cite{zare2024ilsurvey, mao2024dexskills, vezzani2022skillsscheduler}.

\subsection{Vision-Language-Action Models}
Vision-Language-Action (VLA) models have recently emerged as a leading paradigm in robotics, largely driven by advances in large language models (LLMs) \cite{intelligence2025pi05, zhong2025vlasurvey}. These models leverage the powerful generalization capabilities of foundation models to map high-level inputs, such as natural language commands or visual scenes, directly to sequences of actions \cite{zitkovich2023rt, reed2022generalist}. However, the success of VLAs is heavily reliant on access to massive, web-scale datasets with rich language and textual annotations, even when fine-tuning from a pre-trained base.

In the context of a control hierarchy, VLAs typically operate as a high-level meta-controller. They excel at semantic reasoning and decision-making -- determining what skill to execute next (e.g., ``pick up the red block") rather than specifying the fine-grained joint-level movements required to execute that skill \cite{zhong2025vlasurvey}. Our work is complementary to this paradigm, focusing on discrete skill low-level primitive learning and execution, while VLAs focus on mapping language to a sequence of behaviors.

\section{METHODOLOGY}

\begin{figure}[thpb]
  \centering
  \includegraphics[width=1\linewidth]{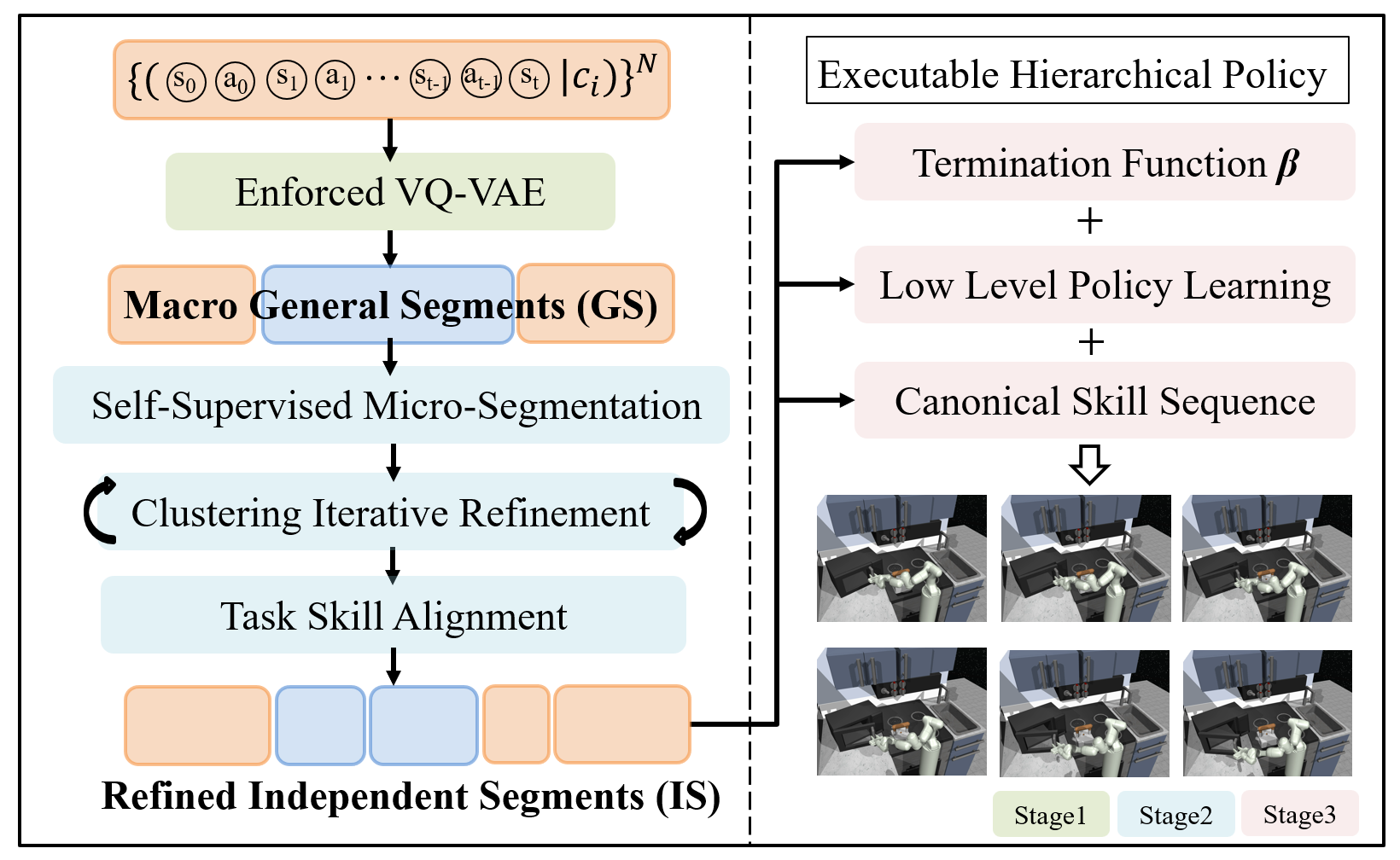}
  \caption{LOKI Framework Diagram. The left panel illustrates the two-stage skill discovery and segmentation process: (light green) Macro-segmentation using the Enforced VQ-VAE, followed by (light blue) Self-Supervised Micro-segmentation with Clustering Iterative Refinement, and Task Skill Alignment, yielding Refined Independent Segments (IS). The right panel depicts the third stage of the framework: constructing executable hierarchical policies from the discovered skills, which includes learning a Termination Function ($\beta$), Low-Level Policy, and deriving a canonical skill sequence policy for task execution. Skill smooth execution is shown as a visual output.}
  \label{fig:loki_diagram}
\end{figure}

In this section, we elaborate on LOKI, our novel three-stage end-to-end learning framework for offline skill discovery and hierarchical imitation. We detail its architecture, learning objectives, and the unique adaptations we have made to existing techniques. The complete workflow of LOKI is visually summarized in Fig. \ref{fig:loki_diagram}, illustrating how raw trajectories are transformed into refined skills that ultimately drive executable hierarchical policies.

\subsection{Stage 1: Macro-Segmentation via Enforced VQ-VAE}
\label{ssec:macro_segmentation}

Let our offline dataset be $\mathcal{D} = \{(\tau_i, c_i)\}_{i=1}^N$, where each trajectory $\tau_i$ consists of a sequence of transitions, and $c_i \in \{1, \dots, M\}$ is a weak label indicating the high-level task to which the trajectory belongs. Each transition is defined as a tuple $x_t = (s_t, a_t, s_{t+1})$.

A standard VQ-VAE \cite{van2017neuralvqvae,DBLP:journals/tmlr/ChenG0M0A24} consists of an encoder $\mathcal{E}$, a decoder $\mathcal{D}$, and a learnable codebook $E = \{e_k\}_{k=1}^K$. The encoder maps an input $x_t$ to a continuous latent embedding $z_e(x_t) = \mathcal{E}(x_t)$. This embedding is then quantized by replacing it with the closest vector from the codebook, $z_q(x_t) = e_{k^*}$, where $k^* = \arg\min_k \| z_e(x_t) - e_k \|_2$. The total loss function for a VQ-VAE is given by:
\begin{equation}
\label{eq:vq_vae_total_loss}
\mathcal{L}_{\text{VQ-VAE}} = \mathcal{L}_{\text{recon}} + \mathcal{L}_{\text{codebook}} + \lambda \mathcal{L}_{\text{commit}}
\end{equation}
Here, the reconstruction loss $\mathcal{L}_{\text{recon}}$ ensures accurate input reconstruction. The codebook loss $\mathcal{L}_{\text{codebook}}$ updates codebook embeddings to match encoder outputs , while the commitment loss $\mathcal{L}_{\text{commit}}$ encourages encoder outputs to commit to specific codebook vectors, preventing codebook fluctuation; $\lambda$ is a weighting hyperparameter for the commitment term.

\subsubsection{Boundary Detection via Latent Space Entropy}
\label{sssec:boundary_detection}
The trained EVQ-VAE provides a principled way to detect skill boundaries. For a given transition $x_t$, we compute the L2 distance between its continuous embedding $z_e(x_t)$ and every codebook vector $e_k$. These distances are then normalized via a softmax function to produce a probability distribution over the codebook:
\begin{equation}
\label{eq:prob_dist}
p_t(k) = \frac{\exp(-\|z_e(x_t) - e_k\|_2^2)}{\sum_{j=1}^K \exp(-\|z_e(x_t) - e_j\|_2^2)}
\end{equation}
We then calculate the entropy of this distribution, $H(p_t) = -\sum_{k=1}^K p_t(k) \log p_t(k)$.

This entropy serves as a powerful signal for segmentation. Extrinsic skills, unique to a specific task, result in a low-entropy distribution, as $z_e(x_t)$ is very close to its task-aligned codebook vector. Conversely, intrinsic skills, common across multiple tasks, produce embeddings roughly equidistant to several codebook vectors, leading to a high-entropy distribution. Significant shifts in this entropy signal along a trajectory indicate change points, marking the boundaries between intrinsic and extrinsic general skills. We use the Pelt algorithm with an RBF kernel from the ruptures library \cite{thurnhofer2020radial} to robustly identify these change points.

\subsubsection{Enforced VQ-VAE (EVQ-VAE)}
\label{sssec:evq-vae_details}

\begin{figure}[t]
  \centering
  \includegraphics[width=1\linewidth]{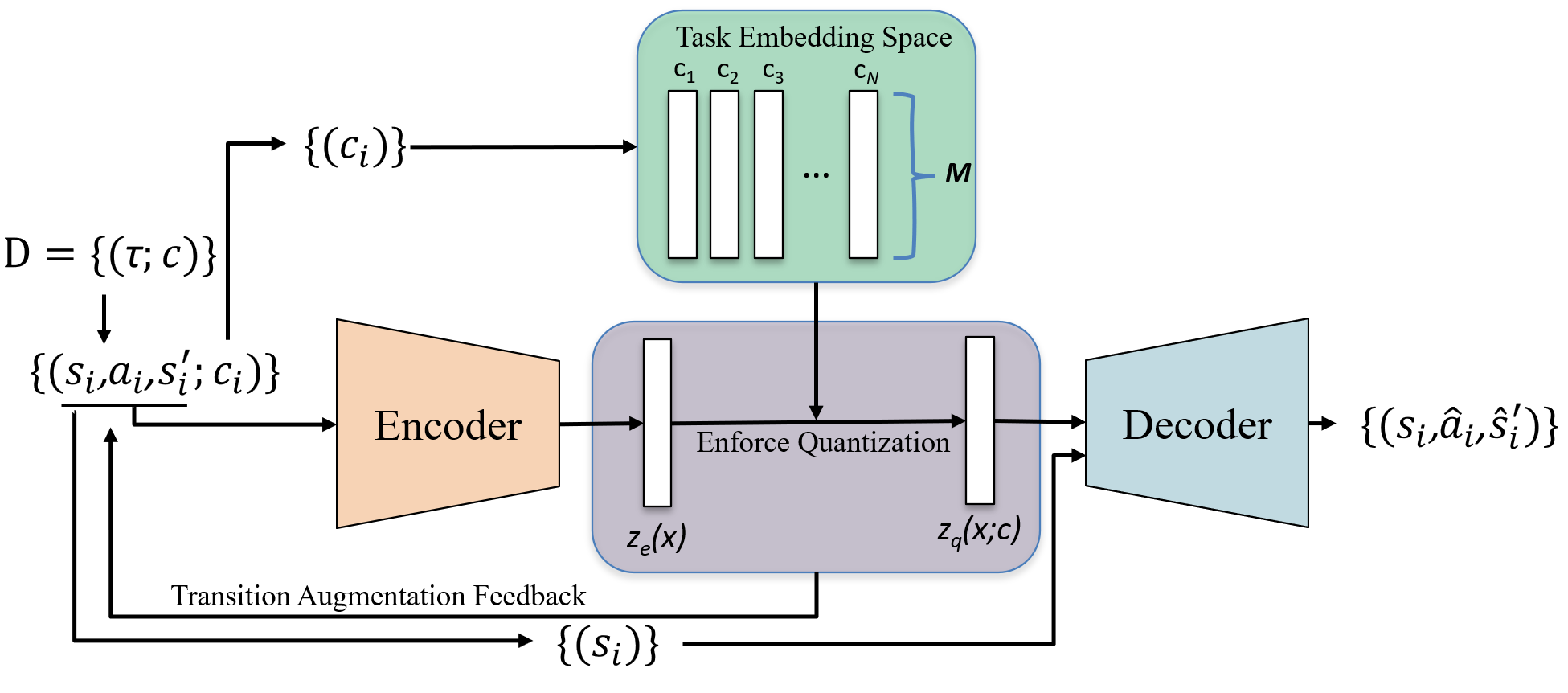}
  \caption{Diagram of the EVQ-VAE architecture that leverages weak task labels $c_i$ to guide an Enforced Quantization from embeddings $z_e(x)$ to codebook vector $z_q(x;c)$. The mechanism achieves an inherent clustering of task-relevant states and actions while maintaining a diverse representation for multi-task behaviors, thereby enabling granular trajectory decomposition. $M$ is the dimension of codebook vector.}
  \label{fig:vq_vae}
\end{figure}

While effective at discretization, a standard VQ-VAE lacks an inherent mechanism to align its latent codes with specific task structures, which is crucial for interpretable skill discovery in multi-task settings. To overcome this limitation and achieve a task-conditional skill learning, we introduce our Enforced Vector Quantized Variational Autoencoder (EVQ-VAE), as depicted in Figure \ref{fig:vq_vae}. Our EVQ-VAE introduces two key modifications: (1) it leverages the weak task label $c$ to enforce a task-conditional quantization, and (2) it incorporates a novel divergence loss to promote disentangled codebook representations.

Given an input transition $x_t=(s_t, a_t, s_{t+1})$ and its associated task label $c$, the EVQ-VAE aims to reconstruct the action and next state, $(\hat{a}_t, \hat{s}_{t+1})$. The total loss function for EVQ-VAE is defined as:
\begin{align}
\label{eq:evq_vae_total_loss_split}
\mathcal{L}_{\text{EVQ-VAE}} = & \mathcal{L}_{\text{recon}}(x;c) + \mathcal{L}_{\text{codebook}}(x;c) \\
             & + \beta \mathcal{L}_{\text{commit}}(x;c) + \gamma \mathcal{L}_{\text{divergence}} \nonumber
\end{align}
The loss terms are detailed as follows.

The task-conditional reconstruction loss ($\mathcal{L}_{\text{recon}}(c)$) encourages the decoder to accurately reconstruct the input $(a_t, s_{t+1})$ from the task-conditional quantized latent representation $z_q(x_t;c)$ and the task label $c$:
\begin{equation}
\label{eq:recon_loss}
\mathcal{L}_{\text{recon}}(x_t;c) = \|a_t - \hat{a}_t\|^2 + \|s_{t+1} - \hat{s}_{t+1}\|^2
\end{equation}
The task-conditional codebook loss $\mathcal{L}_{\text{codebook}}(x_t;c)$ and commitment loss $\mathcal{L}_{\text{commit}}(x_t;c)$ are adapted to ensure that codebook vectors and encoder outputs are learned consistently with task conditioning. Critically, the quantization now explicitly considers the task label $c$. Instead of simply selecting the globally nearest codebook vector, the enforcement mechanism ensures that the chosen $z_q(x_t;c)$ is the nearest codebook vector that is relevant to task $c$. This guidance implicitly directs the encoder's output $z_e(x_t)$ towards task-specific embeddings and updates the codebook accordingly. The specific forms are:
\begin{equation}
\label{eq:codebook_loss_c}
\mathcal{L}_{\text{codebook}}(x_t;c) = \| \text{sg}[z_e(x_t)] - z_q(x_t;c) \|_2^2
\end{equation}
\begin{equation}
\label{eq:commit_loss_c}
\mathcal{L}_{\text{commit}}(x_t;c) = \| z_e(x_t) - \text{sg}[z_q(x_t;c)] \|_2^2
\end{equation}
where $\text{sg}[\cdot]$ is the stop-gradient operator

The codebook divergence loss ($\mathcal{L}_{\text{divergence}}$) is introduced to encourage distinct and specialized representations among the codebook vectors, promoting the discovery of unique extrinsic skills and preventing codebook collapse. It is defined as the negative sum of squared Euclidean distances between all unique pairs of codebook vectors; $\gamma$ is the weighting coefficient for this term:
\begin{equation}
\label{eq:divergence_loss}
\mathcal{L}_{\text{divergence}} = -\sum_{k=1}^K \sum_{l=k+1}^K \|e_k - e_l\|_2^2
\end{equation}

\subsection{Stage 2: Weakly-Supervised Micro-Segmentation}
\label{ssec:micro_segmentation}

Following the macro-segmentation of trajectories into General Skills (GS) using EVQ-VAE, our second stage focuses on further decomposing these macro-segments into finer-grained, context-independent Independent Skills (IS). This micro-segmentation is achieved through a weakly-supervised approach that refines skill boundaries with high precision, building upon the principles of the OPAL framework \cite{ajay2020opal}.

For this purpose, we employ a sequential VAE architecture. Instead of processing entire macro-segments of variable length, we use a sliding window of a fixed size to create training samples. Given a macro-segment, we generate a set of fixed-length windows $\mathcal{D}_{\text{win}} = \{\tau_{\text{win}}\}$, where each window $\tau_{\text{win}} = \{(s_t, a_t, s_{t+1})\}_{t=i}^{i+W-1}$ has a length of $W$. The objective of the model is to precisely reconstruct these windows and to organize its latent space in a manner that aligns with a prior distribution, which is conditioned on the window's initial state and the weak task label. The total loss is defined as:
\begin{equation}
\label{eq:micro_total_loss}
\mathcal{L}_{\text{micro}} = \mathbb{E}_{\tau_{\text{win}} \sim \mathcal{D}_{\text{win}}} \left[ \mathcal{L}_{\text{recon}} + \alpha_{\text{KL}} \mathcal{L}_{\text{KL}} \right]
\end{equation}

The reconstruction loss ($\mathcal{L}_{\text{recon}}$) encourages the model to capture detailed dynamics by reconstructing both the actions $a_t$ and subsequent states $s_{t+1}$ within the window:
\begin{equation}
\label{eq:micro_recon_loss}
\mathcal{L}_{\text{recon}} = \sum_{t=i}^{i+W-1} \left( \|s_{t+1} - \hat{s}_{t+1}\|_2^2 + \|a_t - \hat{a}_t\|_2^2 \right)
\end{equation}
The Kullback-Leibler (KL) divergence term ($\mathcal{L}_{\text{KL}}$) regularizes the posterior distribution $q(z|\tau_{\text{win}})$, which is encoded from the entire window, against a prior distribution $p(z|s_i, c)$. This prior is conditioned only on the window's initial state $s_i$ and the task label $c$. This design is crucial for hierarchical execution, where the policy cannot observe the future. It ensures that the latent code $z$, which represents a skill and summarizes a future sequence of behavior, can be reasonably approximated by a prior that only depends on the current state. The KL divergence is expressed as:
\begin{equation}
\label{eq:micro_kl_loss}
\mathcal{L}_{\text{KL}} = \text{KL}(q(z|\tau_{\text{win}}) || p(z|s_i, c))
\end{equation}

\subsubsection{Skill Stability and Boundary Detection}
\label{sssec:skill_stability}

A key characteristic of distinct skills is their temporal stability. When an agent performs a well-defined skill, our model's ability to reconstruct its state and action transitions is high, resulting in a low reconstruction error. Conversely, a deviation from the current skill's expected dynamics often signals a potential skill change or boundary. We find that a potential skill transition is therefore reliably indicated by a sudden spike in the reconstruction error signal. To robustly identify these spikes, we analyze the local geometry of the error curve. Specifically, we identify candidate split points at time steps corresponding to significant local peaks in the second derivative of the smoothed error signal. These peaks accurately capture points of maximum upward concavity, which correspond to the onset of a ``spike", allowing us to generate an initial set of potential skill boundaries within the macro-segments.

\subsubsection{Iterative Refinement of Micro-Segmentation}
\label{sssec:iterative_refinement}

The initial candidate split points generated from the error signal may still be noisy or overly granular. To obtain functionally distinct independent skills, we employ an iterative refinement process to systematically merge redundant segments. The process begins by dividing a trajectory into smaller segments based on the current set of candidate split points. Each segment is then encoded into a latent embedding using our trained sequential VAE encoder. A K-Means clustering algorithm is subsequently applied to all segment embeddings, assigning a cluster label to each. The crucial refinement step occurs when two consecutive segments within the same trajectory are assigned to the same cluster; the candidate split point separating them is considered redundant and is removed. This iterative merging process is applied exclusively to the candidate split points identified during this micro-segmentation stage, while the ``hard" boundaries established during macro-segmentation are preserved to maintain the overall task structure. The process repeats for a fixed number of rounds or until the set of split points converges.

\subsubsection{Task Skill Alignment}
\label{sssec:task_skill_alignment}

After obtaining a stable set of refined split points, the final step in skill discovery is to assign a meaningful and consistent skill ID to each segment. Our approach employs a Task Skill Alignment strategy that distinguishes between reusable, task-agnostic (intrinsic) skills and unique, task-specific (extrinsic) skills. First, we perform K-Means clustering exclusively on the latent embeddings of all identified intrinsic segments, grouping them into a set of reusable skill primitives $\{k_1, \dots, k_{K_{\text{int}}}\}$. In contrast, each distinct extrinsic skill is directly assigned a unique, non-overlapping skill ID from a separate set $\{k_{K_{\text{int}}+1}, \dots, k_K\}$, treating them as specialized primitives that are not transferrable across tasks. Finally, to enforce structural consistency, we mandate that all trajectories belonging to the same task must share an identical skill sequence. For each task, we determine the most probable or ``canonical" skill sequence by analyzing the similarity and frequency of observed sequences across its demonstrations. All trajectories for that task are then force-aligned to this single canonical sequence. This process yields a final, structured dataset where each segment is labeled with a semantically coherent and consistent skill ID, ready for the final stage of policy learning.

\subsection{Hierarchical Policy Learning}
\label{ssec:hierarchical_policy}

Once the trajectories have been segmented into a set of discrete, independent skills, the final stage of our framework is to construct a hierarchical policy capable of executing these skills to solve the designated tasks. This policy consists of three key components: a learned state-based termination function, a skill-conditioned low-level policy, and a deterministic high-level task program.

\subsubsection{Termination Function Learning}
\label{sssec:termination_function_learning}
To manage the transition between skills, we learn a state-based termination function, $\beta(s) \rightarrow [0, 1]$, adapted from the Option Framework \cite{SUTTON1999semimdp}. Unlike methods that rely on fixed-length time horizons for skills, our approach learns to terminate a skill based on reaching a meaningful state boundary. We formulate this as a binary classification problem, where the final state of each discovered skill segment is labeled as a positive example (indicating termination), and all other states within the skill segments are labeled as negative. A Multi-Layer Perceptron is trained on this data to learn the function $\beta(s)$. During execution, a skill is terminated when the output of $\beta(s_t)$ exceeds a pre-defined threshold.

\subsubsection{Low-Level Policy via Conditional Behavior Cloning}
\label{sssec:low_level_policy_learning}
The execution of each atomic skill is handled by a single, unified low-level policy, $\pi(a_t | s_t, k)$, trained via conditional behavior cloning. The policy is conditioned on the current state $s_t$ and a one-hot encoding of the desired skill ID $k \in \{1, \dots, K\}$. The training dataset is composed of all state-action pairs from the refined skill segments, with each pair $(s_t, a_t)$ labeled with its corresponding skill ID $k$. The policy is then trained with a standard supervised learning objective to maximize the log-likelihood of the demonstrated actions:
\begin{equation}
\label{eq:bc_loss}
\mathcal{L}_{\text{BC}} = \sum_{(s_t, a_t, k) \in \mathcal{D}_{\text{segments}}} -\log \pi(a_t | s_t, k)
\end{equation}
where $\mathcal{D}_{\text{segments}}$ is the comprehensive dataset of all discovered skill segments from Stage 2 as detailed in Section \ref{ssec:micro_segmentation}.

\subsubsection{Canonical Skill Sequences as a High-Level Policy}
\label{sssec:canonical_skill_sequences}
The high-level policy in our framework is not a learned, dynamic controller. Instead, we leverage the highly consistent skill sequences discovered and aligned in Stage 2 (see Section \ref{sssec:task_skill_alignment}) as deterministic, high-level task programs. During execution, the agent simply follows this pre-defined sequence for a given task, transitioning to the next skill upon the completion of the current one. This approach drastically simplifies the learning problem, making the control structure more robust and interpretable by transforming the need for skill selection into a direct execution of a pre-determined script.

\section{EXPERIMENTS}

\subsection{Experimental Setup}
\label{ssec:experimental_setup}

\subsubsection{Environment and Dataset}
\label{sssec:environment_dataset}
\begin{figure}[h!]
    \centering
    % --- First Row ---
    \begin{minipage}{0.15\textwidth}
        \includegraphics[width=\linewidth]{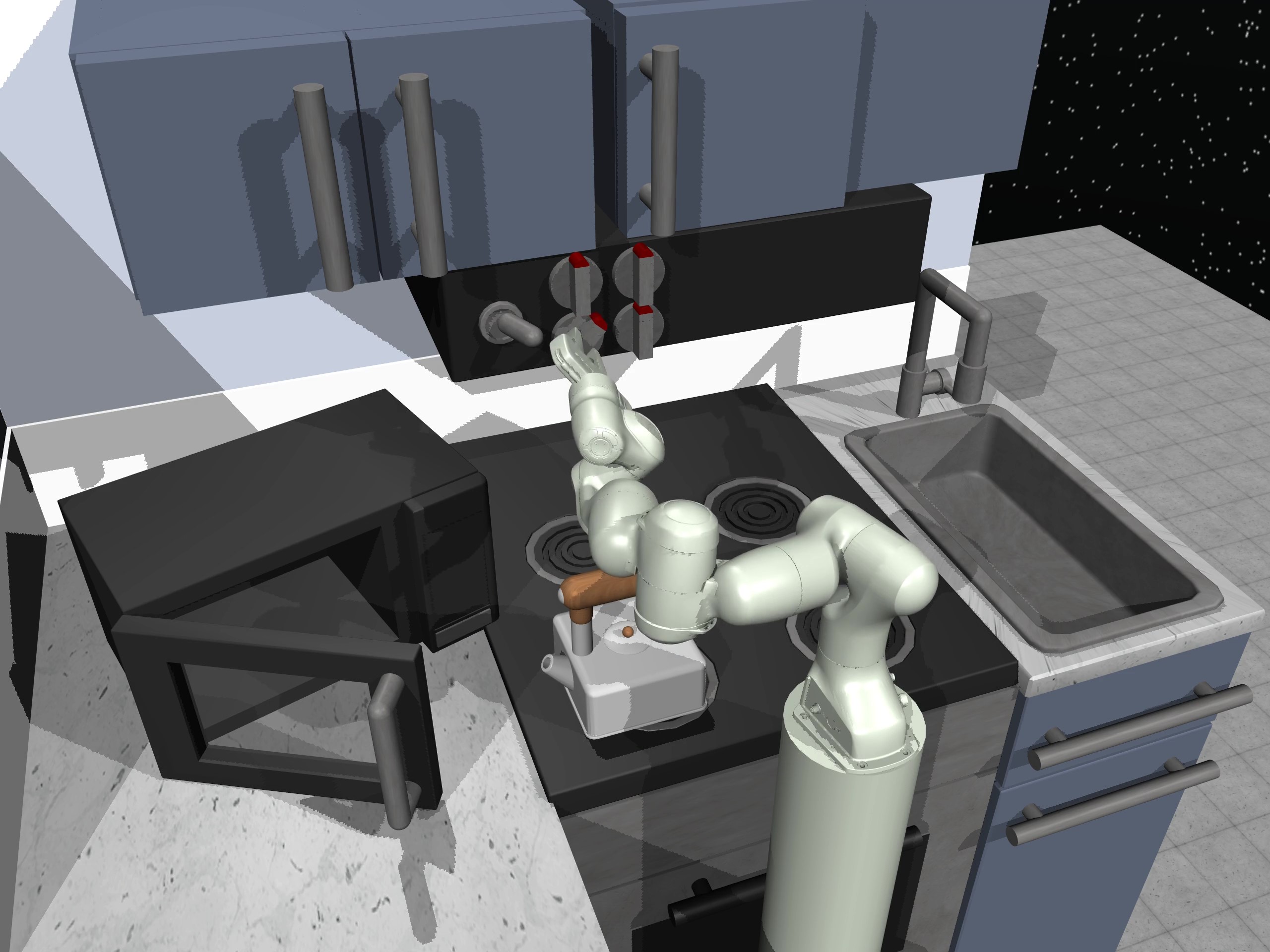}
        \centering
        \footnotesize{(a) Bottom Burner}
    \end{minipage}
    \hfill % Adds horizontal space
    \begin{minipage}{0.15\textwidth}
        \includegraphics[width=\linewidth]{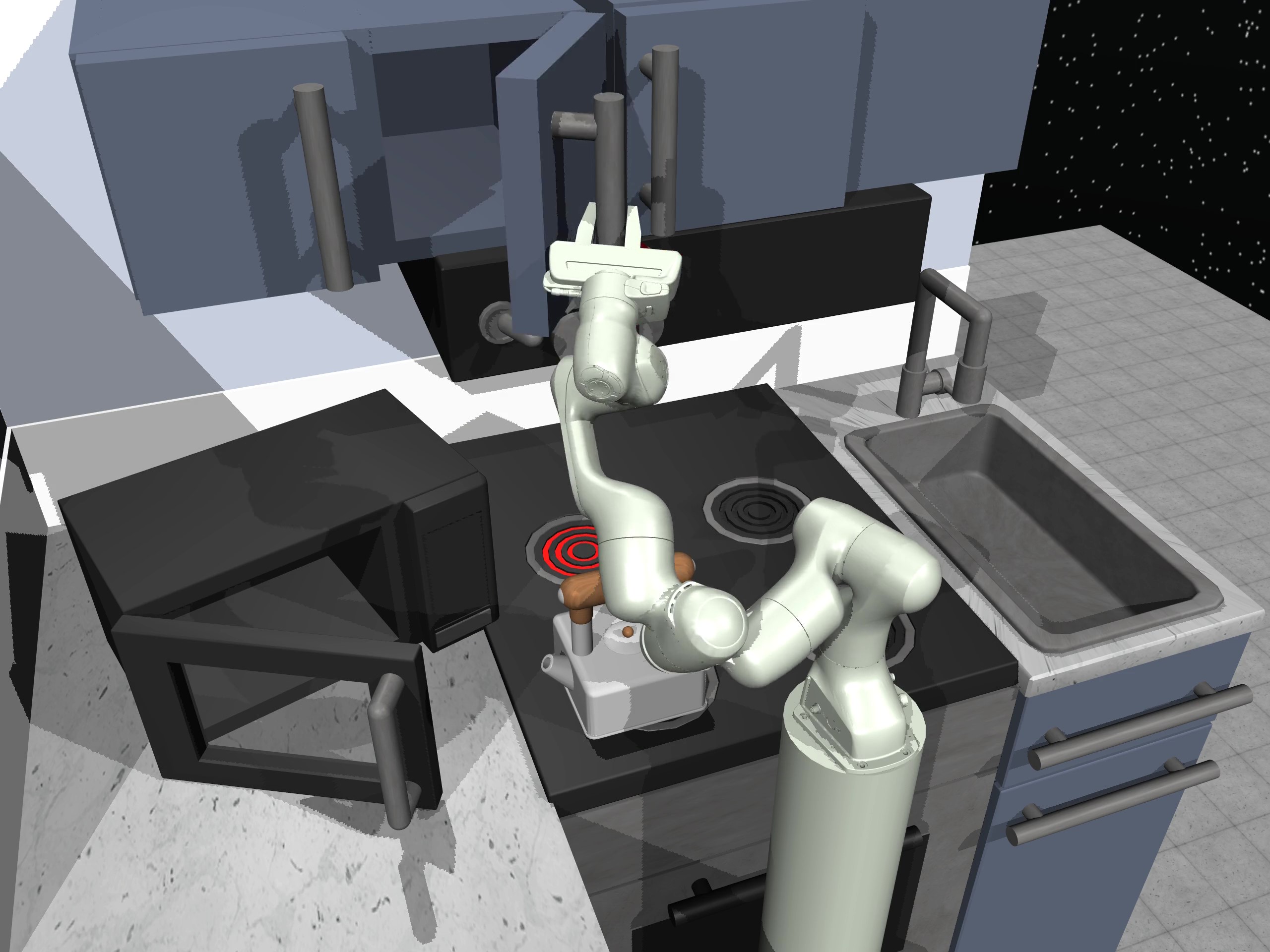}
        \centering
        \footnotesize{(b) Hinge Cabinet}
    \end{minipage}
    \hfill % Adds horizontal space
    \begin{minipage}{0.15\textwidth}
        \includegraphics[width=\linewidth]{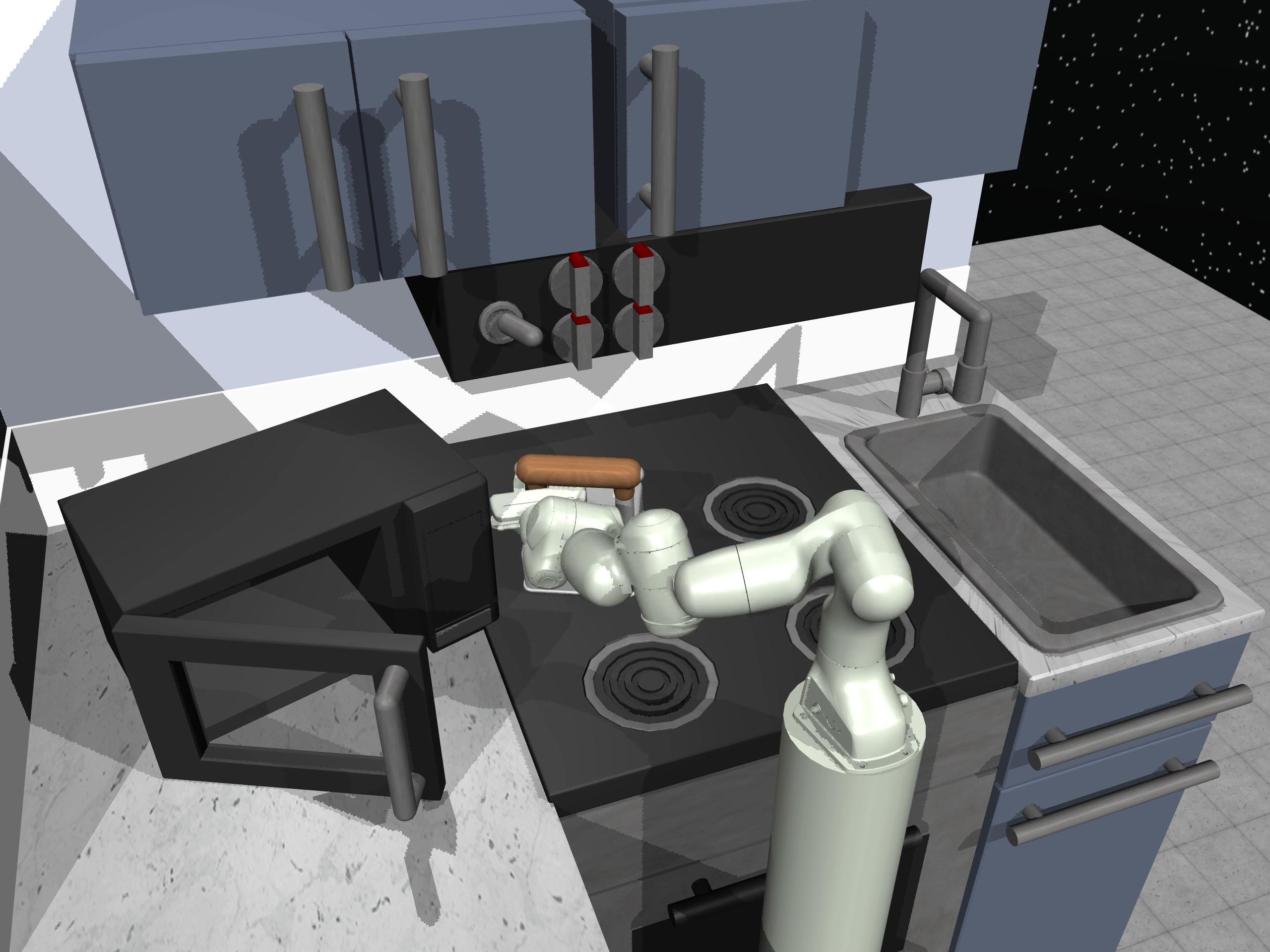}
        \centering
        \footnotesize{(c) Kettle}
    \end{minipage}
    
    \vspace{1em} % Adds some vertical space between the rows

    % --- Second Row ---
    \begin{minipage}{0.15\textwidth}
        \includegraphics[width=\linewidth]{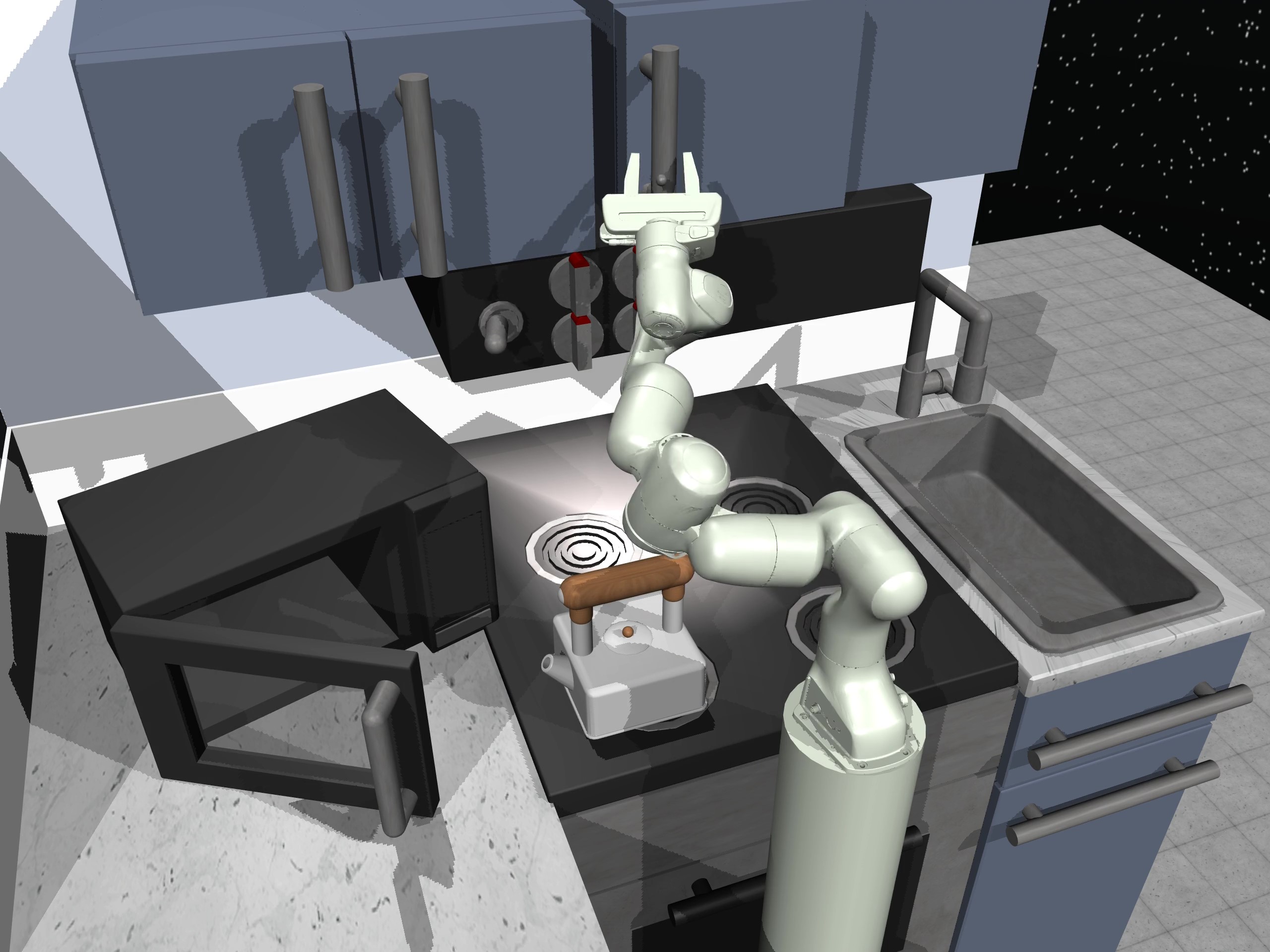}
        \centering
        \footnotesize{(d) Slide Cabinet}
    \end{minipage}
    \hfill % Adds horizontal space
    \begin{minipage}{0.15\textwidth}
        \includegraphics[width=\linewidth]{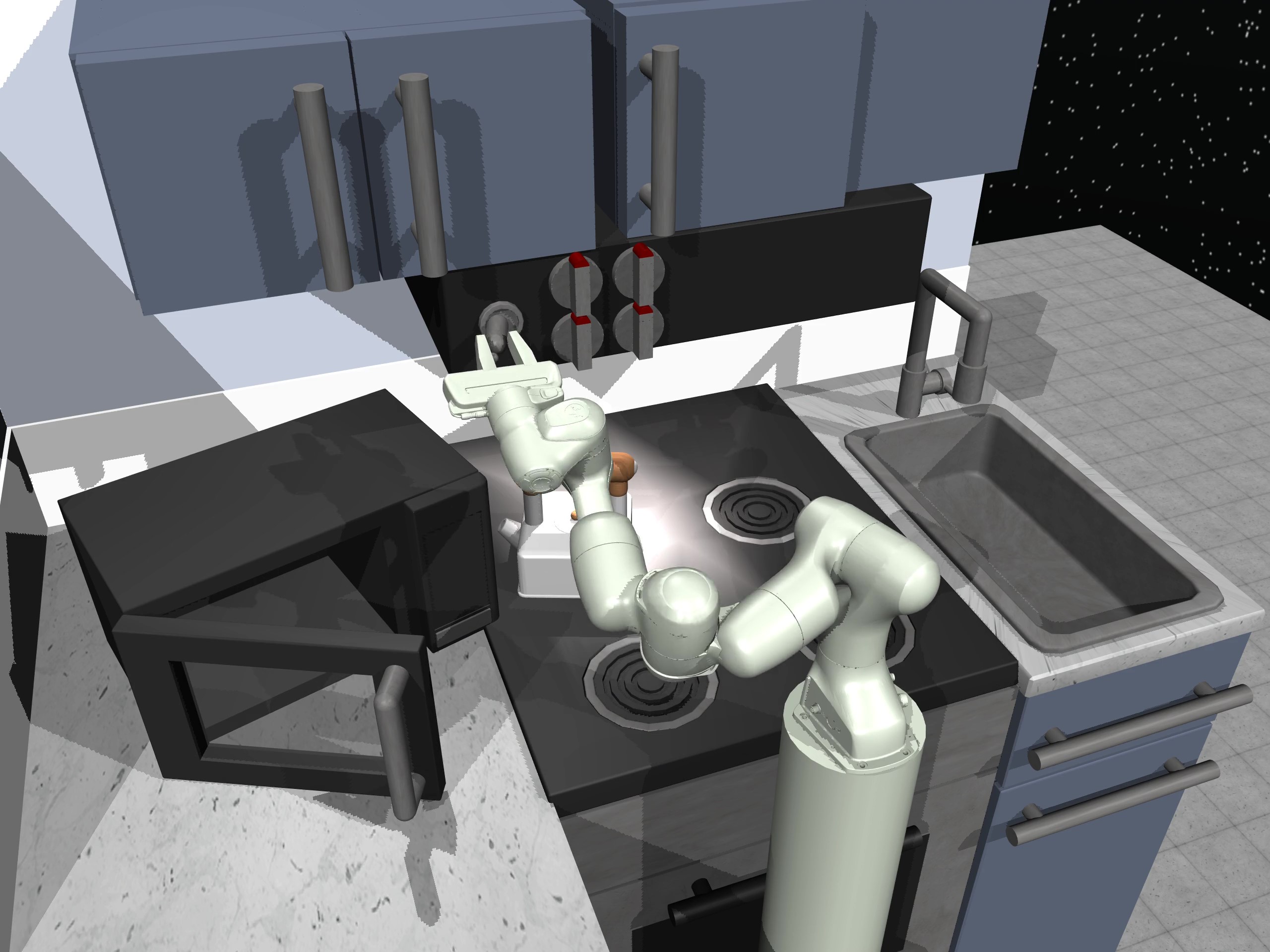}
        \centering
        \footnotesize{(e) Light Switch}
    \end{minipage}
    \hfill % Adds horizontal space
    \begin{minipage}{0.15\textwidth}
        \includegraphics[width=\linewidth]{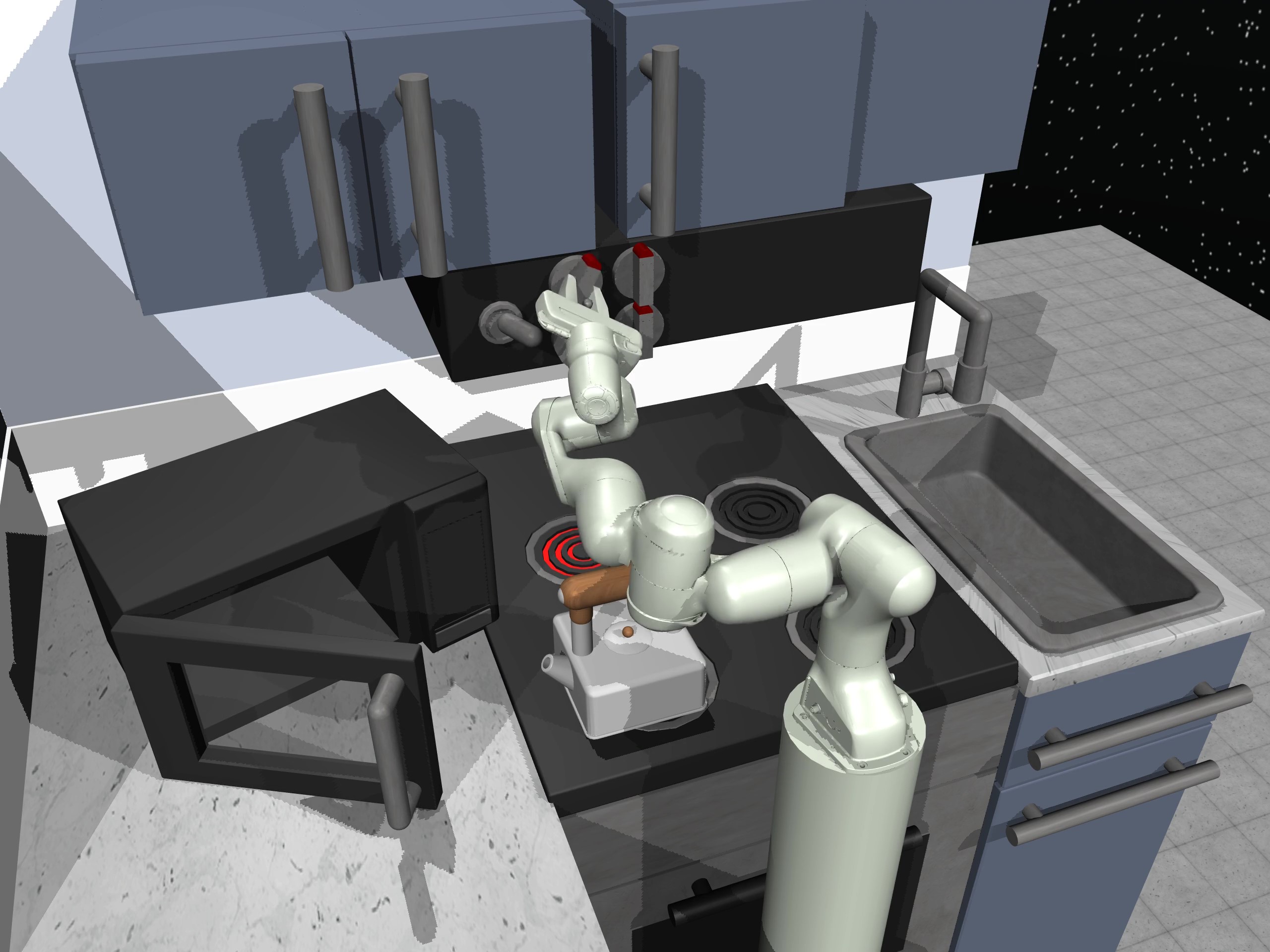}
        \centering
        \footnotesize{(f) Top Burner}
    \end{minipage}

    \vspace{1em} % Adds some vertical space between the rows

    % --- Third Row (Single Image) ---
    \begin{minipage}{0.15\textwidth}
        \includegraphics[width=\linewidth]{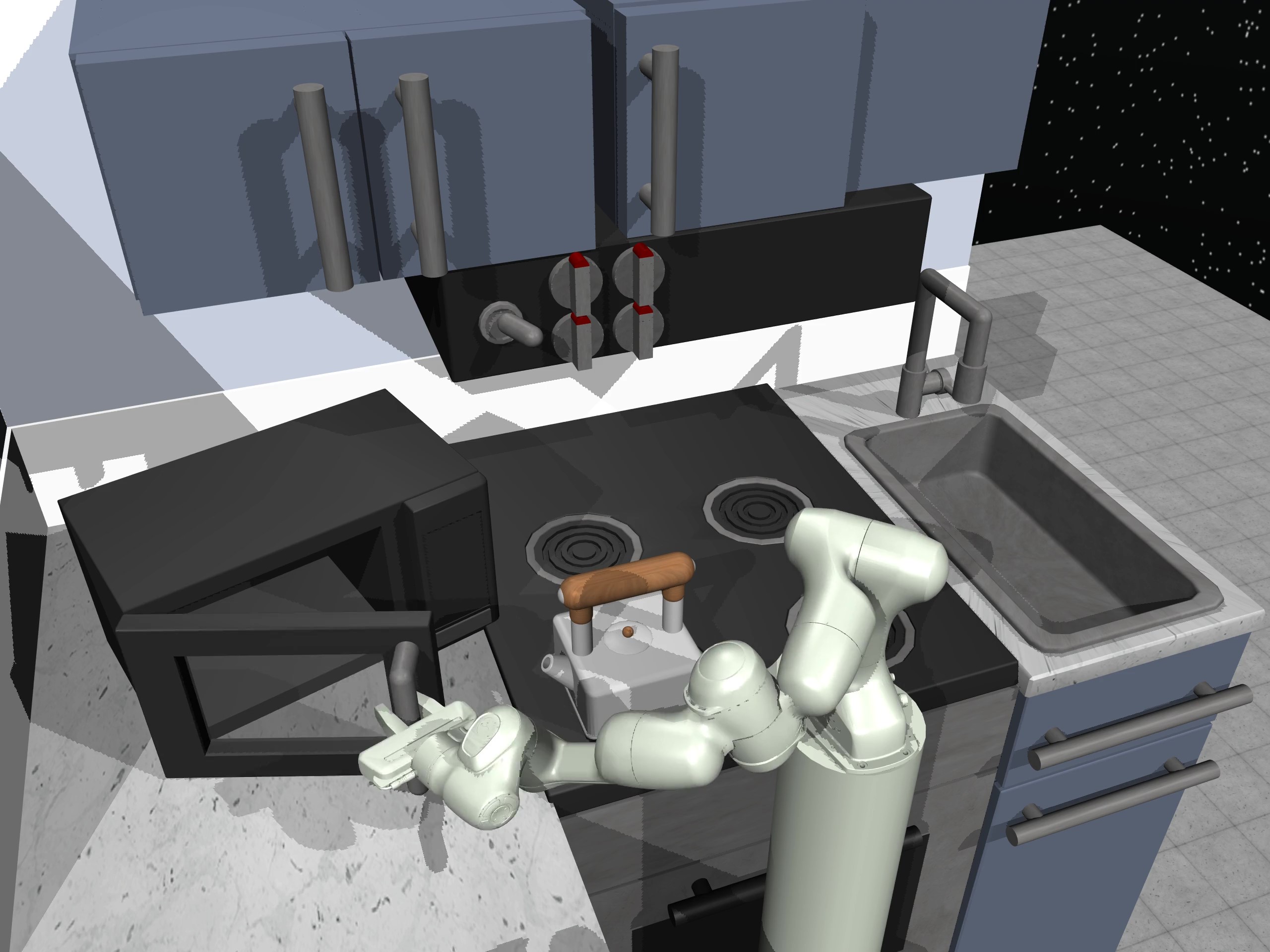}
        \centering
        \footnotesize{(g) Microwave}
    \end{minipage}
    
    % The MAIN caption for the entire figure
    \caption{An overview of the different subtasks in the D4RL Kitchen dataset \cite{fu2020d4rl}. The complete trajectory dataset includes seven subtasks: (a) Turn on Bottom Burner; (b) Open Hinge Cabinet; (c) Move Kettle; (d) Turn on Slide Cabinet; (e) Turn on Light Switch ; (f) Turn on Top Burner; (g) Open Microwave, and each trajectory contains four subtasks.}
    \label{fig:subtasks_overview}
\end{figure}

Our experiments are conducted within the challenging Franka Kitchen manipulation environment, a standard benchmark for long-horizon, multi-task robotic learning. We utilize the `kitchen-mixed-v0' offline dataset from the D4RL benchmark \cite{fu2020d4rl}, generated using the relay policy learning framework \cite{gupta2019relaypolicylearning}. This comprehensive dataset features a rich set of behaviors corresponding to seven distinct subtasks, which include interacting with objects like a microwave, kettle, light switch, and various cabinets. These atomic subtasks are visually represented in Figure \ref{fig:subtasks_overview}. For clarity, we use a representative word to denote each subtask in the subsequent discussions.

\subsubsection{Task Selection and Data Structure}
\label{sssec:task_selection}
From the full D4RL dataset, we select a subset of roughly 24 successful demonstration trajectories for each of three specific complex, multi-stage tasks to evaluate our method. These tasks are as follows and we use initials to represent these task for short:
\begin{itemize}
  \item Microwave Kettle Switch Hinge - MKSH
  \item Microwave Bottom Top Hinge - MBTH
  \item Microwave Switch Slide Hinge - MSSH
\end{itemize}
This refined dataset, comprising 71 successful trajectories in total, forms the basis for our experimental evaluation. Among seven subtasks, Microwave, Hinge, and Switch are intrinsic skills, while Kettle, Bottom, Top, and Slide are extrinsic Skills.

\subsubsection{Data Characteristics and Supervision}
\label{sssec:data_characteristics}
Our offline dataset presents a significant challenge due to its inherent characteristics. Each trajectory consists solely of a sequence of state-action pairs $(s_t, a_t)$, and crucially, contains no reward signals or explicit annotations for subtask boundaries. The demonstrations are collected from successful executions, meaning each trajectory achieves its final task goal. The only form of supervision we provide is a single, weak task label for each trajectory, indicating which of the three complex tasks it belongs to. This minimal supervision setting is designed to rigorously test our model's ability to discover meaningful, reusable skills without requiring extensive manual annotations or dense reward signals.

\subsection{Results and Analysis}

\subsubsection{Macro-Segmentation Analysis}
\begin{figure}[thpb]
 \centering
 \includegraphics[width=1\linewidth]{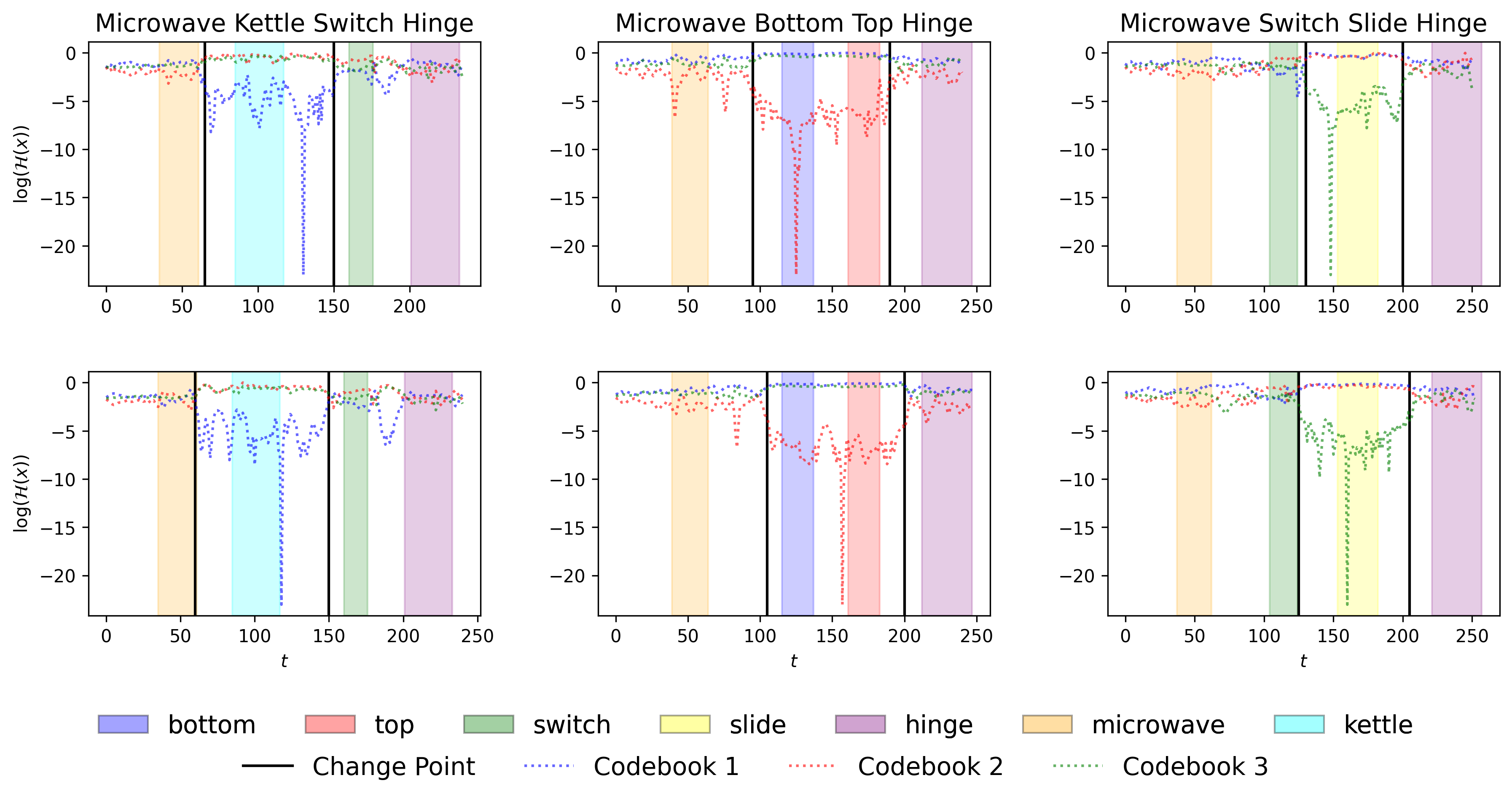}
 \caption{Change point detection results from our EVQ-VAE on three complex tasks. The y-axis represents the log entropy, calculated from the L2 distance between a transition's embedding and each codebook vector. Dotted lines of different colors (blue, red, green) correspond to the log probability of the embedding belonging to each of the three primary codebook vectors. A low-entropy state, where one codebook vector is dominant (one dotted line drops significantly), indicates an extrinsic (task-specific) skill. A high-entropy state, where the lines are tangled and no single codebook is dominant, represents an intrinsic (task-agnostic) skill. The colored shaded regions denote the ground-truth duration of each subtask, while the solid black vertical lines are the change points detected by our method.}
 \label{fig:vq_vae_seg_simple_example}
\end{figure}

Figure \ref{fig:vq_vae_seg_simple_example} illustrates the macro-segmentation results from our Enforced VQ-VAE (EVQ-VAE) on representative trajectories from the three complex tasks. The results demonstrate that our weakly-supervised approach successfully identifies the boundaries between distinct general skills. The plots show the log entropy of the codebook probability distribution over the course of a trajectory. As hypothesized in Section \ref{sssec:boundary_detection}, the entropy signal serves as a robust indicator for distinguishing between intrinsic and extrinsic skills.

We observe a consistent pattern across all tasks. During extrinsic skills, which are unique to a specific task, the transition embeddings are strongly aligned with a single task-specific codebook vector, resulting in a low-entropy state. Conversely, during intrinsic skills that are common across multiple tasks, the embeddings are more equidistant to several codebook vectors, leading to a high-entropy state. For instance, in the MKSH task, the EVQ-VAE correctly identifies the \texttt{kettle} subtask as a low-entropy extrinsic skill, while the surrounding subtasks exhibit the high-entropy signature of intrinsic skills. This clear delineation validates our EVQ-VAE's ability to perform robust macro-segmentation, effectively decomposing complex, unlabeled trajectories into a foundational sequence of General Skills for subsequent micro-level refinement.

\subsubsection{Micro-Segmentation Analysis}
\label{sssec:micro_segmentation_analysis}

\begin{figure}[thpb]
 \centering
 \includegraphics[width=1\linewidth]{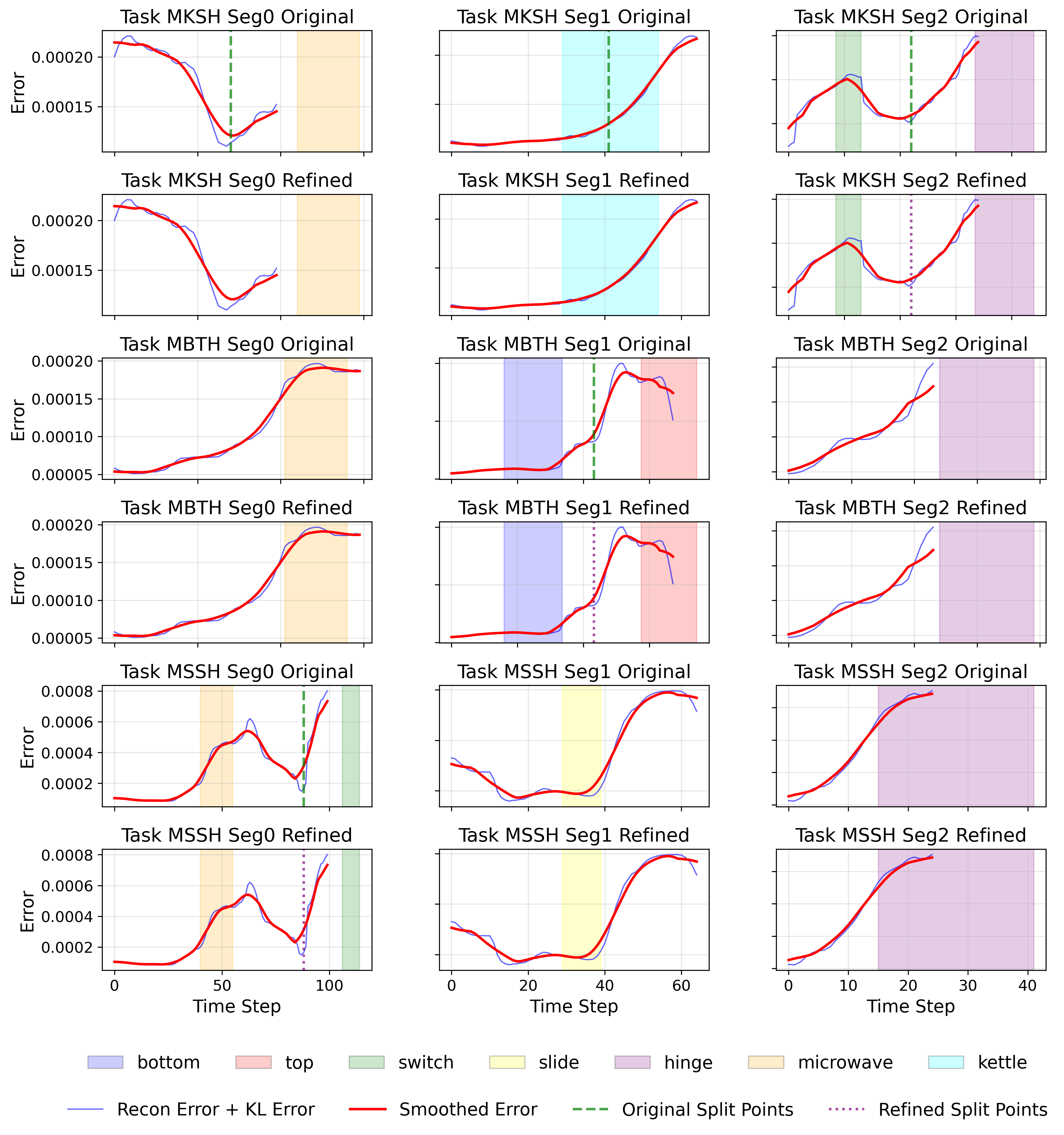}
 \caption{Results of the iterative refinement process on micro-segmentation split points. The figure displays reconstruction error curves for macro-segments from our three evaluation tasks. Odd-numbered rows ('Original') show the initial candidate split points (green dashed lines) detected from the error signal. Even-numbered rows ('Refined') show the final split points (purple dotted lines) after our iterative refinement and task alignment. Each column represents a different macro-segment (Seg0, Seg1, etc.) within its respective task. The refinement process effectively removes noisy, redundant split points while enforcing a consistent segmentation structure across all demonstrations of a given task.}
 \label{fig:trajectory_segmentation_stage2}
\end{figure}

To evaluate the effectiveness of our Stage 2 micro-segmentation process, we analyze the quality of the discovered skill boundaries both before and after our iterative refinement procedure. Figure \ref{fig:trajectory_segmentation_stage2} presents this comparison across representative macro-segments from our three evaluation tasks. The initial split points (green dashed lines), generated from the reconstruction error signal, successfully identify most major skill transitions. However, they can be noisy or overly granular, often placing multiple redundant boundaries within a single, continuous subtask (e.g., ``Task MKSH Seg0 Original").

Our iterative refinement and task alignment mechanism effectively addresses these issues. As shown in the ``Refined" rows, the process successfully removes spurious split points by merging consecutive segments that are assigned to the same skill cluster. For instance, in the ``Task MKSH Seg0 Refined" plot, the multiple initial splits within the `microwave' subtask are correctly consolidated. Furthermore, the task alignment component ensures that all trajectories belonging to the same high-level task converge to a consistent segmentation structure. This process not only removes unnecessary boundaries but can also enforce a necessary split point if it was missed in a particular trajectory but present in others of the same task type. The final refined split points (purple dotted lines) demonstrate a clean, robust, and consistent segmentation that aligns well with the ground-truth subtask durations.

% The table you provided, ready to be used
\begin{table*}[t]
\caption{Comparison of Baseline Performance (Low-Level Policy MSE $\downarrow$). Lower values indicate better imitation fidelity.}
\label{tab:baselines_results_spanning}
\centering
\begin{tabular}{l ccc c}
\toprule
% --- New Two-Row Header ---
\multirow{2}{*}{\textbf{Method}} & \multicolumn{3}{c}{\textbf{Task}} & \multirow{2}{*}{\textbf{Average}} \\
\cmidrule(lr){2-4} % 在第2到4列下画一条短线
& \textbf{MKSH} & \textbf{MBTH} & \textbf{MSSH} & \\
\midrule
% --- Table Data ---
\textbf{LOKI-O (Ours)} & 7.01$\pm$23.6 & 7.23$\pm$11.9 & 6.47$\pm$9.34 & \textbf{6.94$\pm$15.3} \\
\textbf{LOKI-E (Ours)} & 7.10$\pm$7.05 & 8.40$\pm$12.1 & 7.43$\pm$7.70 & \textbf{7.76$\pm$9.69} \\
DADS & 13.5$\pm$20.2 & 12.6$\pm$29.3 & 12.2$\pm$18.4 & 12.9$\pm$23.5 \\
OPAL & 11.6$\pm$24.1 & 12.6$\pm$21.7 & 9.79$\pm$23.0 & 10.6$\pm$22.9 \\
Expert (Flat BC) & 1.91$\pm$3.78 & 2.32$\pm$6.29 & 1.91$\pm$2.58 & 2.09$\pm$4.78 \\
\bottomrule
\end{tabular}
\end{table*}

\subsubsection{Skill Human Knowledge Alignment}
\label{sssec:skill_human_knowledge_alignment}

\begin{figure}[thpb]
 \centering
 \includegraphics[width=1\linewidth]{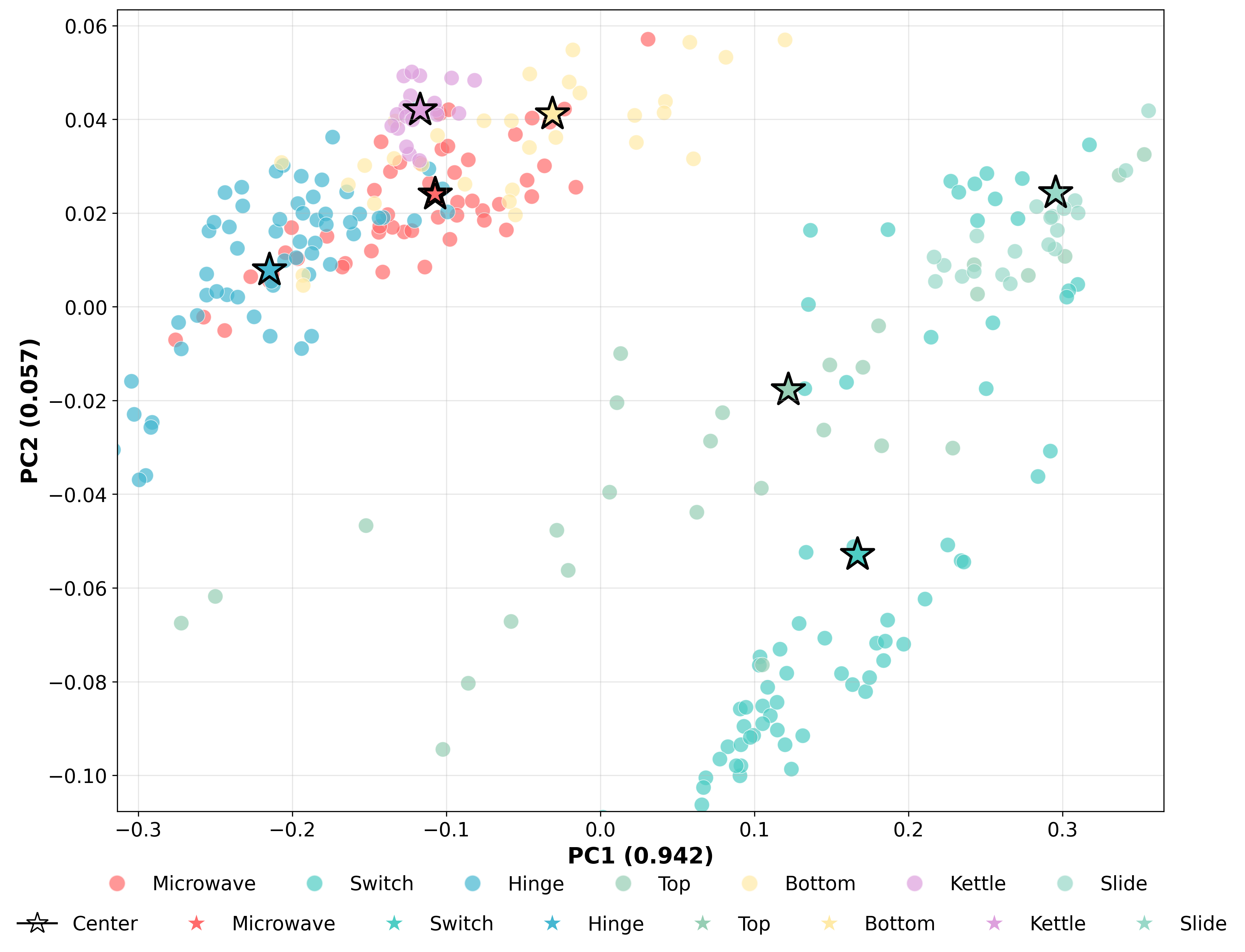}
 \caption{PCA visualization of the discovered skill latent space. Each point is a refined independent skill segment, colored by its ground-truth subtask label. The black-outlined stars indicate the final K-Means cluster centers, representing the learned canonical skills. The plot shows that despite the inherent noise and variance in the raw demonstrations (indicated by the spread of same-colored points), our method successfully identifies distinct cluster centers. The proximity of certain clusters (e.g., Microwave and Hinge) suggests the space has learned a semantically meaningful structure based on motion similarity.}
 \label{fig:final_latent_space_skills}
\end{figure}

To validate the quality and interpretability of the discovered skills, we visualize the latent space of all refined independent skill segments. Figure \ref{fig:final_latent_space_skills} shows a 2D PCA projection of these segment embeddings. The visualization reveals that while the embeddings of segments corresponding to the same ground-truth subtask exhibit considerable variance -- reflecting the natural diversity and noise in human demonstrations -- our clustering process successfully identifies stable and well-separated centroids (black-outlined stars) for each core skill.

Crucially, the learned latent space demonstrates a meaningful semantic structure that aligns with human knowledge. Skills that involve similar robot motions or object interactions are positioned closer to one another. For example, the clusters for 'Microwave' and 'Hinge Cabinet', both of which involve rotational arm movements, occupy adjacent regions in the space. This semantic proximity is highly beneficial for hierarchical control, as it can facilitate smoother and more robust transitions between functionally related skills. This result confirms that our framework not only segments trajectories effectively but also learns a skill representation that is structured, interpretable, and aligned with human conceptualizations of the task.

\subsubsection{Termination Function}
\label{sssec:termination_function_results}

The effectiveness of our hierarchical policy hinges on the ability of the termination function, $\beta(s)$, to accurately signal the completion of a skill. To evaluate this, we measure its performance as a binary classifier on a held-out test set of states. We define a correct prediction not as a single timestep, but as a reasonable interval, $\pm$4 timesteps, around the ground-truth skill boundary to account for minor stochasticity in execution. Our learned termination function achieves a high accuracy of 93.7\% within this defined interval. This result indicates that the skill boundaries discovered during our segmentation stages correspond to meaningful and learnable state transitions, providing a robust and reliable signal for the high-level policy to switch between skills.

\subsubsection{Comparison with Baselines}
\label{sssec:baseline_comparison}

We compare the performance of our method against several baselines using the one-step-ahead prediction Mean Squared Error (MSE) on a held-out test set. This offline metric effectively evaluates the quality of the learned low-level policies and the underlying skill representations they are conditioned on. The results are summarized in Table \ref{tab:baselines_results_spanning}.

The 'Expert' baseline is a flat (non-hierarchical) Behavior Cloning (BC) model conditioned on the high-level task label. Its low MSE represents a strong performance on in-distribution imitation but lacks the modularity and interpretability of a skill-based approach. Our proposed method, LOKI, is presented in two variants: LOKI-O, which uses one-hot skill IDs for conditioning, and LOKI-E, which uses the continuous skill embeddings from the sequential VAE's latent space.

As shown in the table, both LOKI variants significantly outperform the hierarchical baselines, DADS and OPAL, demonstrating the superiority of our two-stage segmentation and task alignment process in discovering a more consistent and imitable set of skills. Notably, LOKI-E achieves a lower MSE compared to LOKI-O. We attribute this to the richer information contained within the continuous skill embeddings. While one-hot IDs treat each skill as an isolated category, the continuous embeddings capture the nuanced variations within a skill and the semantic relationships between different skills. This richer conditioning signal enables the low-level policy to generate more precise and contextually appropriate actions. Overall, these results validate that LOKI effectively balances high-fidelity imitation with the discovery of a structured, interpretable, and reusable skill vocabulary.

% \subsection{Ablation Comparison}

% Why not use the one-hot method in EVQ-VAE?

% why not transformer to encode rather than bi-GRU? small dataset

% recover ability comparing to human split

% \begin{table}[h]
% \caption{Comparison of Baseline Performance (Metric: e.g., Low-Level MSE $\downarrow$). Lower values indicate better performance. Results are reported as mean $\pm$ standard deviation over multiple runs.}
% \label{tab:baselines_results_transposed}
% \begin{center}
% \setlength{\tabcolsep}{5pt} % 稍微减小列间距以适应
% \begin{tabular}{|l||c|c|c||c|}
% \hline
% \textbf{Method} & \textbf{MKSH} & \textbf{MBTH} & \textbf{MSSH} & \textbf{Average} \\
% \hline
% \hline
% LOKI-O & 7.01$\pm$23.6 & 7.23$\pm$11.9 & 6.47$\pm$9.34 & 6.94$\pm$15.3 \\
% LOKI-E & XX.X$\pm$X.X & XX.X$\pm$X.X & XX.X$\pm$X.X & XX.X$\pm$X.X \\
% DADS & 13.5$\pm$20.2 & 12.6$\pm$29.3 & 12.2$\pm$18.4 & 12.9$\pm$23.5 \\
% OPAL & 11.6$\pm$24.1 & 12.6$\pm$21.7 & 9.79$\pm$23.0 & 10.6$\pm$22.9 \\
% Expert & 1.91$\pm$3.78 & 2.32$\pm$6.29 & 1.91$\pm$2.58 & 2.09$\pm$4.78 \\
% \hline
% \end{tabular}
% \end{center}
% \end{table}

\section{Conclusion and Future Work}
\label{sec:conclusion}

In this paper, we addressed the challenge of discovering reusable skills from complex, multi-task offline demonstrations that lack explicit rewards or subtask labels. We introduced LOKI, a three-stage framework that tackles this problem through a novel, weakly-supervised segmentation process. This process first identifies coarse, task-aware behaviors and subsequently refines them into fine-grained, independent skills. Our experiments, conducted in the challenging D4RL Kitchen environment, demonstrated that this approach is highly effective. The discovered skills were not only semantically meaningful and strongly aligned with human intuition, but also enabled a hierarchical policy to achieve high success rates in task reproduction. Ultimately, our work presents a practical and robust method for extracting structured, interpretable skills from raw offline data, offering a significant step towards more scalable and generalizable robot learning from demonstrations.

For future work, we have identified several promising directions. First, we plan to apply and evaluate our framework in more complex and diverse robotic environments to further test its scalability and robustness. A second critical extension is to adapt our framework to handle high-dimensional, image-based inputs instead of structured state observations, which would significantly broaden its applicability. Finally, we aim to validate the generalizability of LOKI's principles on other standard offline skill discovery benchmarks to further establish its effectiveness as a versatile tool for robot learning from demonstration.

\addtolength{\textheight}{-12cm}   % This command serves to balance the column lengths

\bibliographystyle{IEEEtran} 
\bibliography{reference}

\end{document}